\documentclass[a4paper,conference]{IEEEtran}
\usepackage{graphicx}

\usepackage{hyperref}
\usepackage{graphicx}
\usepackage{caption}
\usepackage{subcaption}
\usepackage{float}
\usepackage{multirow}
\usepackage{xcolor}
\usepackage[ruled]{algorithm2e}
\usepackage[export]{adjustbox}
\usepackage{soul}
\usepackage{booktabs}
\usepackage{comment}
\usepackage{amssymb}
\usepackage{amsthm}
\usepackage{amsmath}

\newcommand{\argmax}[1]{\underset{#1}{\operatorname{argmax}}\;}

\usepackage{cite}
\usepackage{footnote}
\usepackage[inline]{enumitem}
\usepackage{threeparttable}

\emergencystretch=\maxdimen
\usepackage{color}

\begin{document}


\title{ScanSSD: Scanning Single Shot Detector for Mathematical Formulas in PDF Document Images}

\author{\IEEEauthorblockN{Parag Mali, Puneeth Kukkadapu, Mahshad Mahdavi, and, Richard Zanibbi}
\IEEEauthorblockA{Department of Computer Science, Rochester Institute of Technology\\Rochester, NY 14623, USA\\
Email: \{psm2208, pxk8301, mxm7832, rxzvcs\}@rit.edu}}






\maketitle
\begin{abstract}

We introduce the Scanning Single Shot Detector (ScanSSD) for locating 
math formulas offset from text and embedded in textlines. ScanSSD uses only visual features for detection: no formatting or typesetting information such as layout, font, or character labels are employed. Given a 600 dpi document page image, a Single Shot Detector (SSD) locates formulas at multiple scales using sliding windows, after which candidate detections are pooled to obtain page-level results. For our experiments we use 
the TFD-ICDAR2019v2 dataset, 
a modification of the GTDB scanned math article collection. 
ScanSSD detects characters in formulas with high accuracy, obtaining a 0.926 f-score, and detects formulas with high recall overall. Detection errors are largely minor, such as splitting formulas at large whitespace gaps (e.g., for variable constraints) and merging formulas on adjacent textlines. Formula detection f-scores of 0.796 (IOU $\geq0.5$) and 0.733 (IOU $\ge 0.75$) are obtained.
Our data, evaluation tools, and code are publicly available. 
\end{abstract}

\IEEEpeerreviewmaketitle

\section{Introduction}
\label{sec:introduction}

The PDF format is used ubiquitously for sharing and printing documents. Unfortunately,
while the latest PDF specification supports embedding structural information for graphical elements (e.g., figures, tables, and footnotes\footnote{\url{https://www.iso.org/standard/63534.html}}), most born-digital PDF documents contain only rendering-level information such as characters, lines, and images. These low-level objects can be recovered by parsing the document \cite{davila2019}, but graphic regions must be located using detection algorithms.
For example, PDFFigures \cite{clark2016pdffigures} extracts figures and tables from born-digital Computer Science research papers for Semantic Scholar.\footnote{\url{https://www.semanticscholar.org}} Unfortunately, older PDF documents may contain only scanned images of document pages, providing {\it no} information about characters or other graphical objects in the page images.

\begin{figure}
	\centering
	\includegraphics[width=0.9\linewidth,fbox]{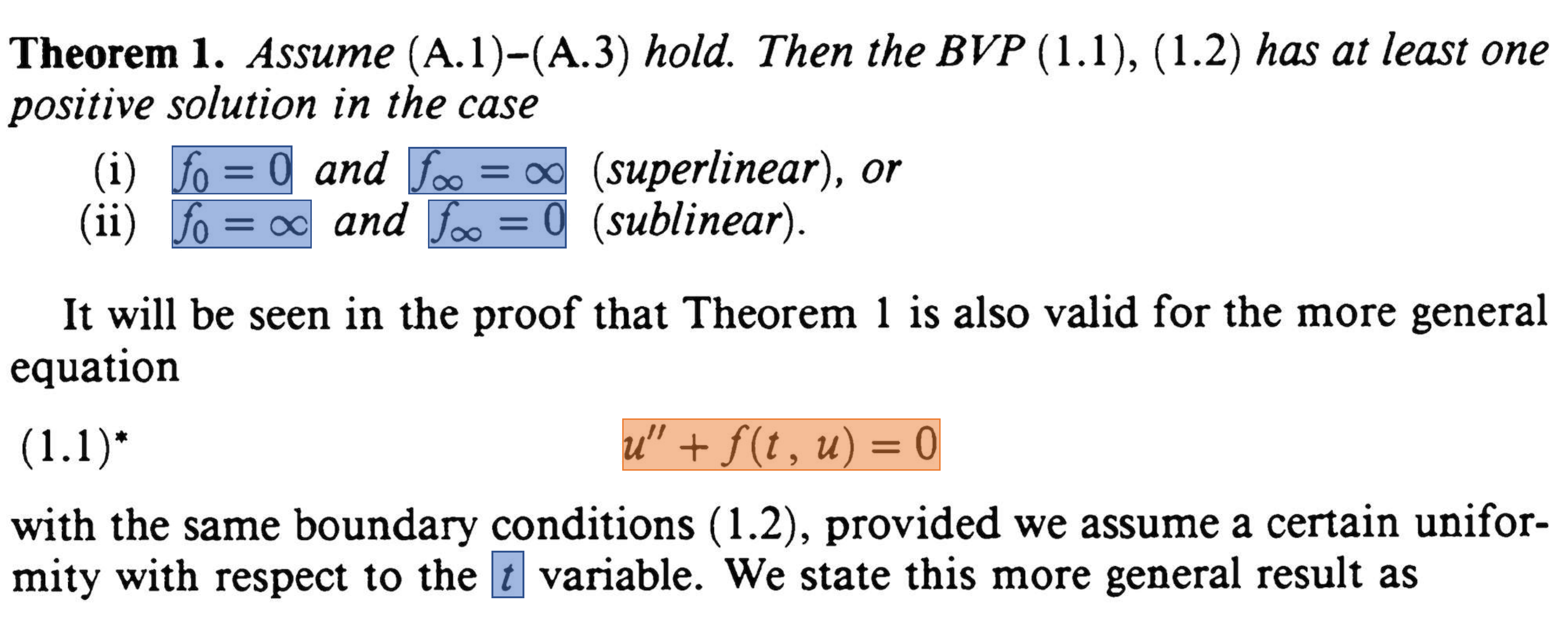}
	\caption[Embedded and displayed math expressions]{Embedded (blue) vs. displayed (red) formulas.}
	\label{fig:embeddeddisplayed}
\end{figure}

We present a new image-based detector for mathematical formulas in in both born-digital and scanned PDF documents.
Math expressions may be \emph{displayed} and offset from the main text, or appear \emph{embedded} directly in text lines (see Figure \ref{fig:embeddeddisplayed}). Displayed expressions are generally easier to detect due to indentation and vertical gaps, whereas embedded expressions are more challenging. Embedded equations may differ in font, e.g., for the italicized variable $t$  in  Figure \ref{fig:embeddeddisplayed}, but italicized words and variations in text fonts make fonts unreliable for detecting embedded formulas in general. 



Our work makes two main contributions. First, we introduce the ScanSSD architecture for detecting formulas using only visual features.
A deep neural-network Single Shot Detector (SSD  \cite{liu2016ssd}) locates formulas at multiple scales using a sliding window in a 600 dpi page image. Page-level formula detections are obtained by pooling SSD region detections using a simple voting and thresholding procedure.
ScanSSD detects characters in formulas with high accuracy (92.6\% f-score),  
and detects formula regions accurately enough to be used as a baseline for indexing mathematical formulas in born-digital and scanned PDF documents.
The ScanSSD code is publicly available.\footnote{\url{https://github.com/MaliParag/ScanSSD}}

Our second contribution is a new benchmark for formula detection comprised of a dataset and evaluation tools. The dataset is a modification of the GTDB database of Suzuki et al \cite{ohyama2019detecting}. 
 Our data and evaluation tools were developed for the ICDAR 2019 TFD competition \cite{Mahdavi2019icdar2019}, and the dataset (TFD-ICDAR2019v2) and evaluation tools are publicly available (see Section \ref{sec:dataset}).
 
 In the next section we provide an overview of related work, followed by our dataset (Section \ref{sec:dataset}), ScanSSD (Sections \ref{sec:window} and \ref{sec:pooling}), our results (Section \ref{sec:results}), and finally our conclusions and plans for future work (Section \ref{sec:conclusion}).


\section{Related Work}
\label{sec:related_work}

Existing methods for formula detection in PDF documents use formatting information, such as page layout, character labels, character locations, font sizes, etc. However, PDF documents are generated by many different tools, and the quality of their character information varies. Lin et al. \cite{lin2011mathematical} point out math formulas may be composed of several object types (e.g. text, image, graph). For example, the square root sign in a PDF generated from \LaTeX{} contains the text object representing a radical sign and a graphical object for the horizontal line. As a result, some symbols must be identified from multiple drawing elements.  

Given characters and formula locations, the visual structure of each formula (i.e., spatial arrangement of symbols on writing lines) can be recovered with high accuracy using existing techniques \cite{Mahdavi2019ls,condonMSc,deng2016image,zhang2018track,alvaro2013shape,zhang2017watch}. 
For formula retrieval, flexible matching of sub-expressions requires that formula structure (i.e., visual syntax and/or semantics) be available - however, there has been recent work using CNN-based embeddings for purely appearance-based retrieval \cite{pfahler2019}.

 Displayed expression detection is relatively easy, as offset formulas differ in height and width of the line, character size, and, symbol layout \cite{garain2007ocr}. Embedded mathematical expressions are more challenging: Iwatsuki et al. \cite{iwatsuki2017detecting} conclude that distinguishing dictionary words that appear in italics and embedded mathematical expressions is a non-trivial task as embedded formulas at times can contain complex mathematical structures such as summations or integrals. However, many embedded math expressions are very small, often just a single symbol in a definition such as `where $w$ is the set of words'. Some approaches have been proposed specifically for embedded math expression detection \cite{lin2012identification, iwatsuki2017detecting} and others specifically for displayed math expressions \cite{drake2005distinguishing, gao2017deep}. 

Lin et al. classify formula detection methods into three categories based on the features used \cite{lin2011mathematical}. These categories are character-based, image-based, and layout-based. Character-based methods use OCR engines to identify characters, and characters not recognized by the engine are considered candidates for math expression elements. The second category of methods uses image segmentation. Most traditional methods require segmentation thresholds. Setting  threshold values can be difficult, especially for the unknown documents. Layout-based methods detect math expressions using features such as line height, line spacing, alignment, etc. Many published methods use a combination of character features, layout features, and context features.

\subsection{Traditional Methods}

Garain and Chaudhari did a survey of over 10,000 document pages and found  the frequency of each mathematical character in formulas \cite{chaudhuri1998approach}. They used this information found to develop a detector for embedded mathematical expressions \cite{garain2000syntactic}. They scan each text line and decide if the line contains one of the 25 most frequent mathematical symbols. After finding the leftmost word containing a mathematical symbol, they grow the region around the word on the left and  right using rules to identify the formula region. For detection of  displayed expressions they use two features: first, white space around  math expressions. Second, the standard deviation of the left lowermost pixels of symbols on the text line. They base this feature on the observation that for a math expression, the leftmost pixels of each symbol are often not on the same line, while for text they often are. A disadvantage of their method for embedded formula detection is that it requires symbol recognition, which adds complexity to the system.  Another approach based on locating mathematical symbols and then growing formula regions around symbols was proposed by Kacem et al., but using fuzzy logic \cite{kacem2001automatic}.

Lin et al. \cite{lin2011mathematical} proposed a four-step detection process. In the first step, they extract the locations, bounding boxes, baselines, fonts, etc., and use them for character and layout features in the following steps. They also process math symbols comprised of multiple objects: for example, a vertical delimiter may be made up of multiple short vertical line objects. They detect named mathematical functions such as `sin,' `cos,' etc., and numbers. In the next step, they distinguish text lines from non-text lines. They find displayed math expressions in non-text lines using geometric layout features (e.g., line-height), character features (e.g., is it the character part of a named math function like `sin'), and context features (e.g., whether the preceding and the following character is a math element). In the last step, they classify characters into math and non-math characters. They find embedded math expressions by merging characters tagged as math characters. SVM classification was used for both isolated math expression detection, and character classification into math and non-math. 

\subsection{CRF and Deep Learning-Based Techniques}

For born-digital PDF papers, Iwatsuki et al. \cite{iwatsuki2017detecting} created a manually annotated dataset and applied conditional random fields (CRF) for math-zone identification using both layout features (e.g. font types) and linguistic features (e.g. n-grams) extracted from PDF documents. For each word, they used three labels: the beginning of a math expression, inside a math expression, and at the end of a math expression.  
They concluded that words and fonts are important for distinguishing math from the text. This method has limitations, as it requires a specially annotated dataset that has each word annotated with either beginning, inside or end of the math expression label. Their method works only for born-digital PDF documents with layout information.    

Gao et al. \cite{gao2017deep} used a combination of CNN and RNN for formula detection. They first extract text, graph and image streams from the PDF document. Next, they perform top-down layout analysis based on XY-cutting \cite{nagy84}, and bottom-up layout analysis based on connected components to generate candidate expression regions. Features are then extracted using neural networks from each candidate region, and they classify candidate regions. Finally, they adjust and refine the incomplete math expression areas. Similar to their method, we use a CNN model (VGG16 \cite{simonyan2014very}) for feature extraction. In contrast to their method, we do not depend on the layout analysis of the page.

Recently Ohyama et al. \cite{ohyama2019detecting} used a U-net to detect characters in formulas. The U-net acts as a pixel-level image filter, and does not produce regions for symbols or formulas. Detection is evaluated based on pixel-level agreement between detected and ground-truth symbols; formula detection is estimated based on the number of formulas with at least half of their characters detected. In contrast, our method produces bounding boxes for mathematical expressions of one or more symbols.  As we wanted to propose specific regions (bounding boxes) for formulas, we decided to explore modern object detection methods employing deep neural networks. 

We next we discuss different object detection methods, and our selection of SSD as the underlying detector for our model. 

\subsection{Object Detection}

The first deep learning algorithm that achieved noticably stronger results for object detection task was the R-CNN \cite{girshick2014rich} (Region proposal with CNN). Unlike R-CNN, which feeds $\approx 2k$ regions to a CNN for each image, in Fast R-CNN\cite{girshick2015fast} only the original input image is used as input. Faster R-CNN \cite{ren2015faster} introduced a different architecture, the \emph{region-proposal network.} In contrast to R-CNN, Fast R-CNN, and Faster R-CNN which use region proposals, the YOLO \cite{redmon2016you} and Single Shot MultiBox Detectors (SSD) \cite{liu2016ssd} perform detection in a single-stage network. Both YOLO and SSD divide the input image into a grid, where each grid point has an associated set of `default' bounding boxes. Unlike YOLO, SSD uses multiple grids with different scales instead of a single grid. This allows an SSD detector to divide the responsibility for detecting objects across scales. The SSD network learns to predict  offsets and size modifications for each default bounding box.  Just like R-CNN, SSD uses the VGG16\cite{simonyan2014very} architecture for feature extraction. SSD does not require selective search, region proposals, or multi-stage networks like R-CNN, Fast R-CNN, and, Faster R-CNN.

Among the CNN-based object detectors, SSD is a simple single stage model that obtains accuracy comparable to models with region proposal steps such as Faster R-CNN \cite{liu2016ssd,huang2017speed}. Liao et al. with their TextBoxes architecture have shown that a modified SSD can detect wide regions \cite{liao2017textboxes}. Formulas are often quite wide, 
and so we use an SSD modified in a manner similar to TextBoxes as the basis for our formula detector. Details for our detector are presented in Section \ref{sec:window}. 

\section{Creating the TFD-ICDAR2019v2 Dataset}
\label{sec:dataset}

For typeset formula detection, we modified ground truth for the GTDB1 and GTDB2 datasets\footnote{available from \url{https://github.com/uchidalab/GTDB-Dataset}} created by Suzuki et al. \cite{suzuki2003infty}. {\it TFD-ICDAR2019v2} represents `Typeset Formula Detection task for ICDAR 2019,' version 2. The first version was used for the CROHME math recognition competition at ICDAR 2019 \cite{Mahdavi2019icdar2019}: version two (v2) adds formulas to ground truth that were missing in the original.
The  dataset is available online, and we provide scripts to compile and render the dataset PDFs at 600 dpi, and evaluation scripts with region matching using thresholded intersection-over-union (IOU) measures.\footnote{ \url{https://github.com/MaliParag/TFD-ICDAR2019} }

The GTDB collection provides annotations for 48 PDF documents from scientific journals and textbooks using a variety of font faces and notation styles. It also provides ground truth at the character level in CSV format, including spatial relationships between math characters (e.g., subscript, superscript). 
Character labels, and an indication of whether a character belongs to a formula region are also provided. 

At the time we created our dataset in early 2019, we were unable to locate two PDFs from GTDB1, and so omitted them in TFD-ICDAR2019v2.\footnote{MA\_1970\_26\_38, and MA\_1977\_275\_292} From the remaining 46 documents, 10 PDFs from GTDB2 serve as the test set (see Figure \ref{fig:filewise}). 
We developed image processing tools for modifying the GTDB ground-truth to reflect scale and translation differences found in the publicly available versions of the PDF documents. GTDB also does not provide bounding boxes for math expressions directly: we used character bounding boxes and spatial relationships to generate math regions in our ground truth files.

Metrics for TFD-ICDAR2019v2 may be found in Table \ref{table:icdar2019}. 
It is worth noting that over 25\% of formulas in the collection contain a single symbol (e.g., `$\lambda$').

\begin{table}[!tbp]
\centering
\caption{TFD-ICDAR2019v2 Collection Statistics.} 
\small
\begin{tabular}{ l  r |    r r r}
\toprule
\multicolumn{1}{l}{} &
\multicolumn{1}{r}{} &
\multicolumn{3}{c}{\textbf{Formulas}}\\

\multicolumn{1}{l}{} &
\multicolumn{1}{r}{\textbf{Docs (Pages)}} &
\multicolumn{1}{r}{1 symbol}&
\multicolumn{1}{r}{$>$1 symbol}&
\multicolumn{1}{r}{Total}\\

\midrule
Training & 36 (569) & 7506 & 18947 & 26453\\
Test & 10 (236) & 2556 & 9350 & 11906\\
\bottomrule
\end{tabular}
\label{table:icdar2019}
\end{table}

\begin{figure*}[!tb]
	\centering
	\includegraphics[width=0.875\linewidth]{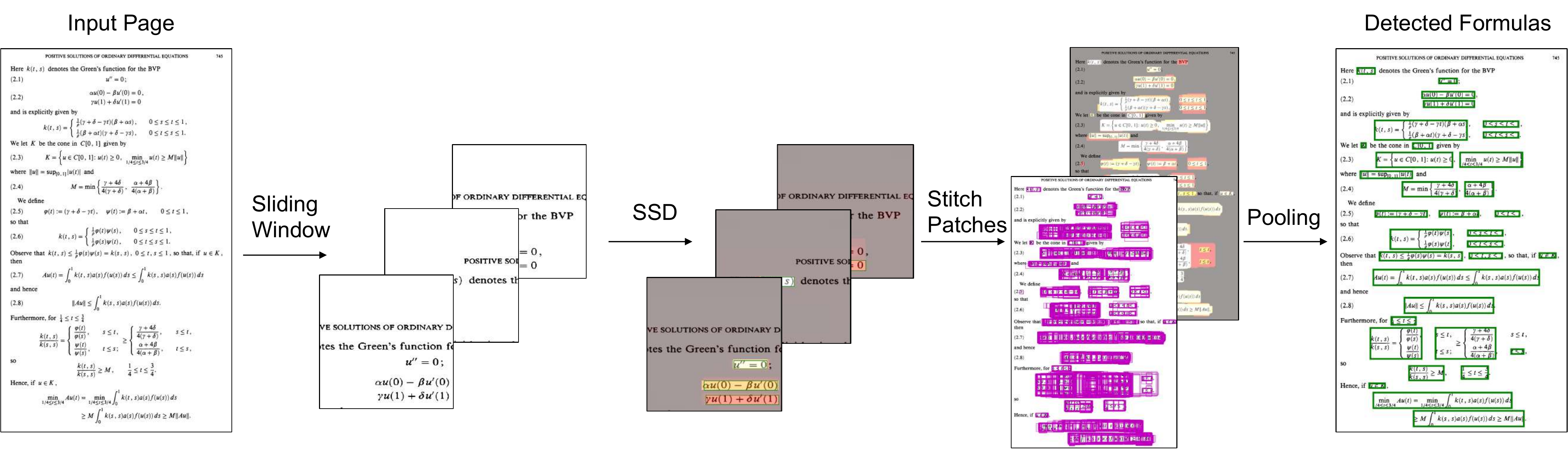}
	\caption[Overview of ScanSSD architecture]{ScanSSD architecture. Heatmaps illustrate detection confidences with  $gray \approx 0$, $red \approx 0.5$, $white \approx 1.0$. Purple and green bounding boxes show formula regions after stitching window-level detections and pooling, respectively.}
	\label{fig:detection_flow}
\end{figure*}

\section{ScanSSD: Window-Level Detection}
\label{sec:window}

\begin{figure}[!b]
    \centering
    \begin{subfigure}[b]{0.40\textwidth}
        \centering
        \includegraphics[width=0.8\textwidth]{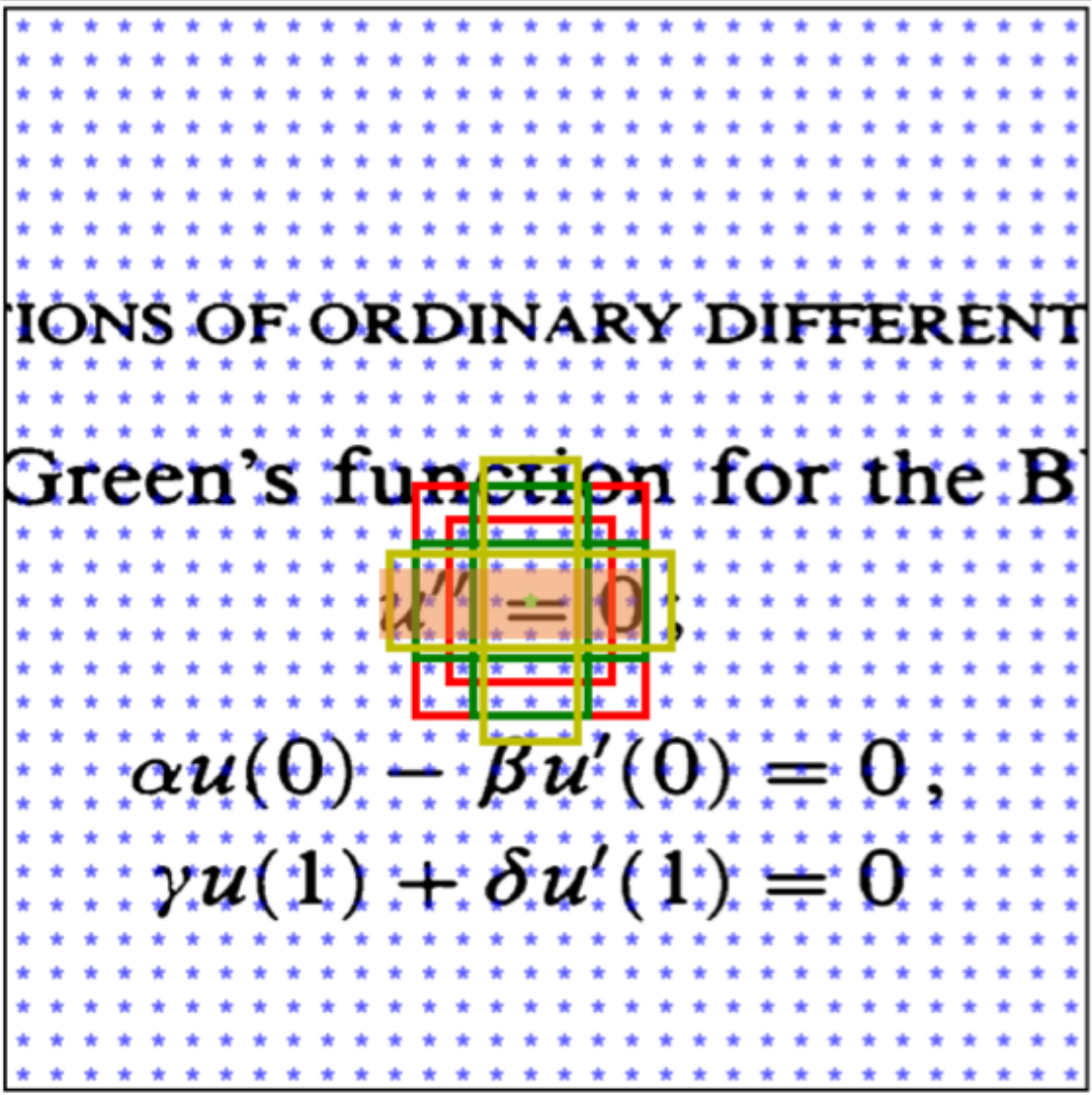}
        \label{fig:default_32}
    \end{subfigure}
    
    \caption[Default boxes for a $512\times 512$ window]{
    Default boxes for a $512\times512$ window. Box centers ($\star$) are in a 32$\times$32 grid. Shown are six default boxes around one point with different sizes and aspect ratios (red, green, and yellow boxes) located near a target formula  (pink highlight).}
    \label{fig:default_boxes}
 \end{figure}

Figure \ref{fig:detection_flow} illustrates
 the ScanSSD architecture. First,  we use a sliding window to sample overlapping sub-images from the document page image. We then pass each window to a Single-Shot Detector (SSD \cite{liu2016ssd}) to locate formula regions.  SSD simultaneously evaluates multiple formula region candidates laid out in a grid (see Figure \ref{fig:default_boxes}), and then applies non-maximal suppression (NMS) to select the window-level detections. NMS is a greedy strategy that keeps one detection per group of overlapping detections.
Formulas detected within each window have associated confidences, shown using colour in the 3rd stage of Figure \ref{fig:detection_flow}.

As seen with the purple boxes in Figure \ref{fig:detection_flow}, many formulas are repeated and/or split across the sampled windows. To obtain page-level formula detections, we first stitch the window-level SSD detections together on the page.  A voting-based pooling method in then used to obtain final detection results (shown as green boxes in Figure \ref{fig:detection_flow}).

Details of the ScanSSD system are provided below.

\subsection{Sliding Windows}

To produce sub-images for use in detection, starting from a 600 dpi page image we slide a $1200\times1200$ window with a vertical and horizontal stride (shift) of 120 pixels (10\% of window size). Our windows are roughly 10 text lines in height, which makes math formulas large enough for SSD to detect them reliably. The SSD detector is trained using ground truth math regions cropped at the boundary of each window, after scaling and translating  formula bounding boxes appropriately. 


{\bf Advantages.} There are four main advantages to using sliding windows. The first  is data augmentation: only 569 page images are available in the training set, which is \emph{very} small for training a deep neural network. Our sliding windows produce 656,717 sub-images. Second, converting the original page image directly to $300 \times 300$ or $512\times512$ loses a great deal of visual information, and when we tried to detect formulas using subsampled page images recall was extremely low.  
Third, as we maintain the overlap between windows, the network sees formulas multiple times, and has multiple chances to detect a formula. This helps increase recall, because formulas appear in more regions of detection windows. Finally, Liu et al. \cite{liu2016ssd} mention that SSD is challenged when detecting small objects. Formulas with just one or two characters are common, but also small. Using high-resolution sub-images increases the relative size of math regions, which makes it easier for SSD to detect them. 

{\bf Disadvantages.} There are also a few disadvantages to using sliding windows versus detection within a single page image. The first is increased computational cost; this can be mitigated through parallelization, as each window may be processed independently.  Secondly, windowing cuts formulas if they do not fit in a window. This means that a large expression may be split into multiple sub-images; this makes it impossible to train the SSD network to detect large math expressions directly. To mitigate this issue, we train the network to detect formulas across windows. Furthermore, windowing requires that we stitch (combine) results from individual windows to obtain  detection results at the level of the original page. We discuss how we address these problems using pooling methods in section \ref{sec:pooling}. 

\subsection{Region Matching and Default Boxes in SSD}
\label{sec:matching_strategy}

SSD defines a fixed space of candidate detection regions organized in a spatial grid at multiple resolutions (`default boxes'). Each default box may be resized and translated by the SSD network to fit target regions, and is associated with a confidence score.
Figure \ref{fig:default_boxes} shows default boxes of different sizes and aspect ratios overlaid on a $512 \times 512$ image. In SSD, each feature map is a pixel grid, but the associated default boxes are defined in the original image coordinate space. The image is analyzed at multiple scales; here for illustration the  $32\times 32$ grid of default boxes is shown. In practice, if we used only the $32 \times 32$ default boxes, we might miss smaller objects. For the highlighted formula in Figure \ref{fig:default_boxes}, the wider yellow box has the maximum intersection-over-union (IOU), and during training the wide yellow box will be matched with the highlighted ground truth.

Our metric for matching ground truth to candidate detection regions is the same as SSD \cite{liu2016ssd}. Each ground truth box is matched to a default box with the highest IOU, and also with default boxes with an IOU greater than 0.5.
Matching targets to more than one default box simplifies learning by allowing the network to predict higher scores for more boxes. The matched default boxes are considered positive examples (POS) and the remaining default boxes are considered negative examples (NEG).

\begin{figure}[b]
	\centering
	\includegraphics[width=0.95\linewidth]{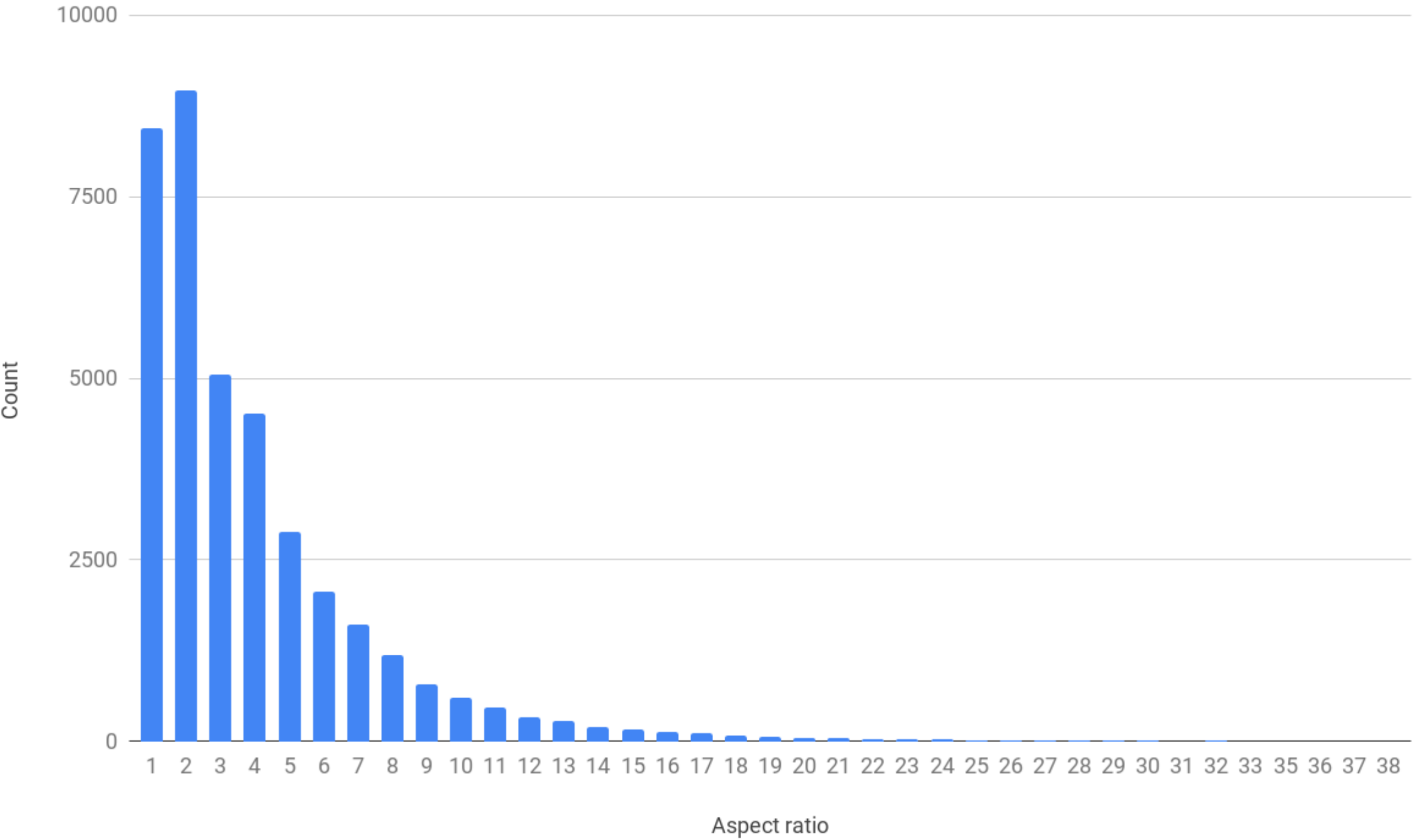}
	\caption{Formula aspect ratios. Most formulas in our dataset have more width than height, i.e., are oriented horizontally.}
	\label{fig:aspect_ratios}
\end{figure}

The original SSD \cite{liu2016ssd} architecture uses aspect ratios (width/height) of $\{1, 2, 3, 1/2, 1/3\}$. However, as we see in Figure \ref{fig:aspect_ratios}, there are many wide formulas with an aspect ratio greater than $3$ in the dataset. As a result, wider default boxes will have a higher chance of matching wide formulas. So, in addition to the default boxes used in the original SSD, we also add the wider default boxes used in TextBoxes\cite{liao2017textboxes}, with aspect ratios $\{5,7,10\}$.  In our early experiments, these wider default boxes increased recall for large formulas.

  
\subsection{Postprocessing}
\begin{figure}
		\centering
		\begin{subfigure}{0.35\textwidth}
        	\includegraphics[width=\linewidth,fbox]{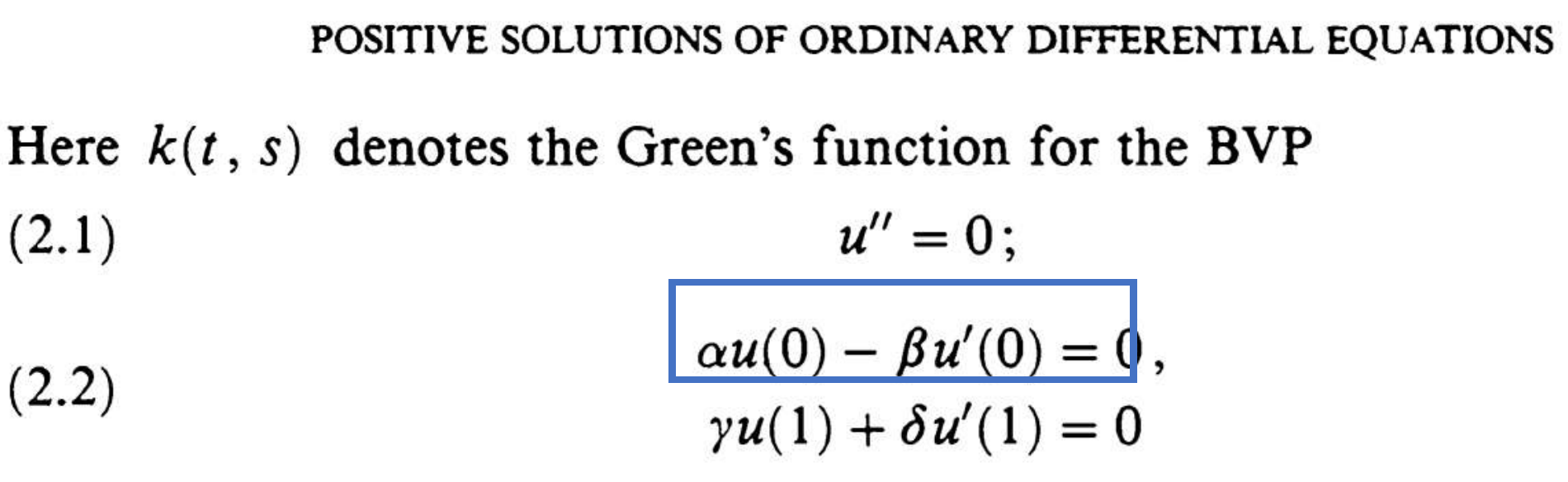}
			\caption{\it Initial detection\\~\\}
		\end{subfigure}
		\hfill
		\begin{subfigure}{0.35\textwidth}
			\includegraphics[width=\linewidth,fbox]{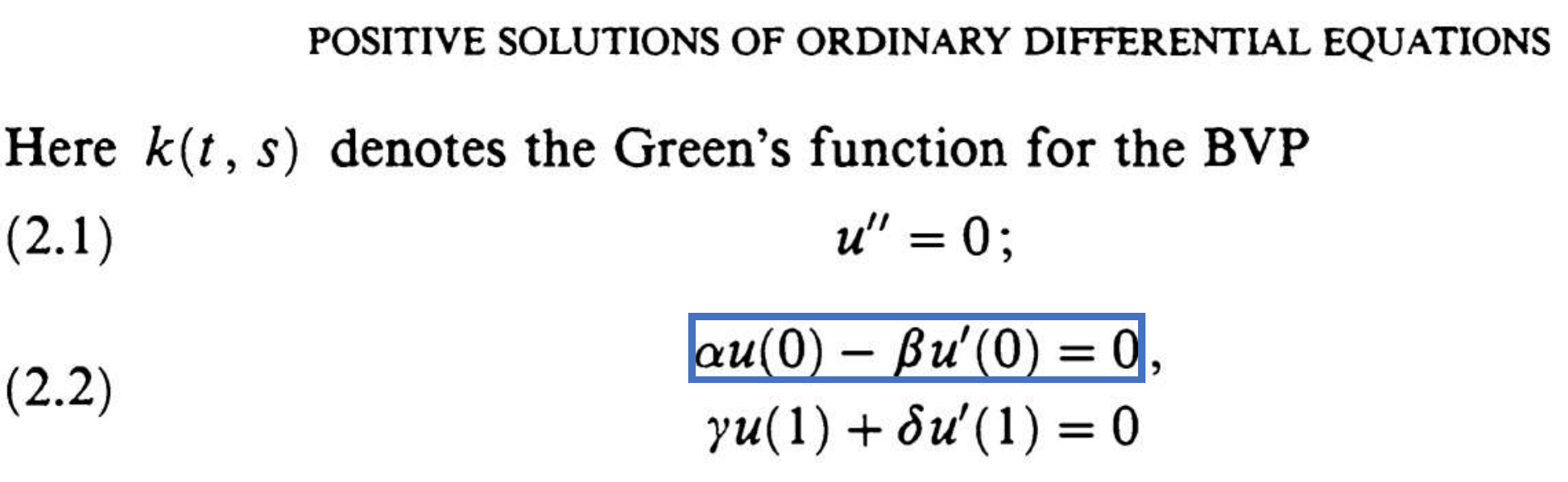}
			\caption{\it After cropping}
		\end{subfigure}
		\caption[Postprocessing]{Postprocessing crops detection regions around connected components within or touching the initial detection.}
		\label{fig:postprocessing}
	\end{figure}

Figure \ref{fig:postprocessing} illustrates postprocessing in ScanSSD. We expand and/or shrink initial formula detections so that are cropped around the connected components they contain and touch at their border. The goal is to capture entire characters belonging to a detection region, without additional padding. This postprocessing is done at two stages: first, before stitching, and second, after pooling regions to obtain output formula detections.
	
\section{ScanSSD: Voting-based Pooling from Windows}
\label{sec:pooling}

\begin{figure*}
	\centering
	\begin{tabular}{cccc}
		\includegraphics[width=0.23\textwidth,fbox]{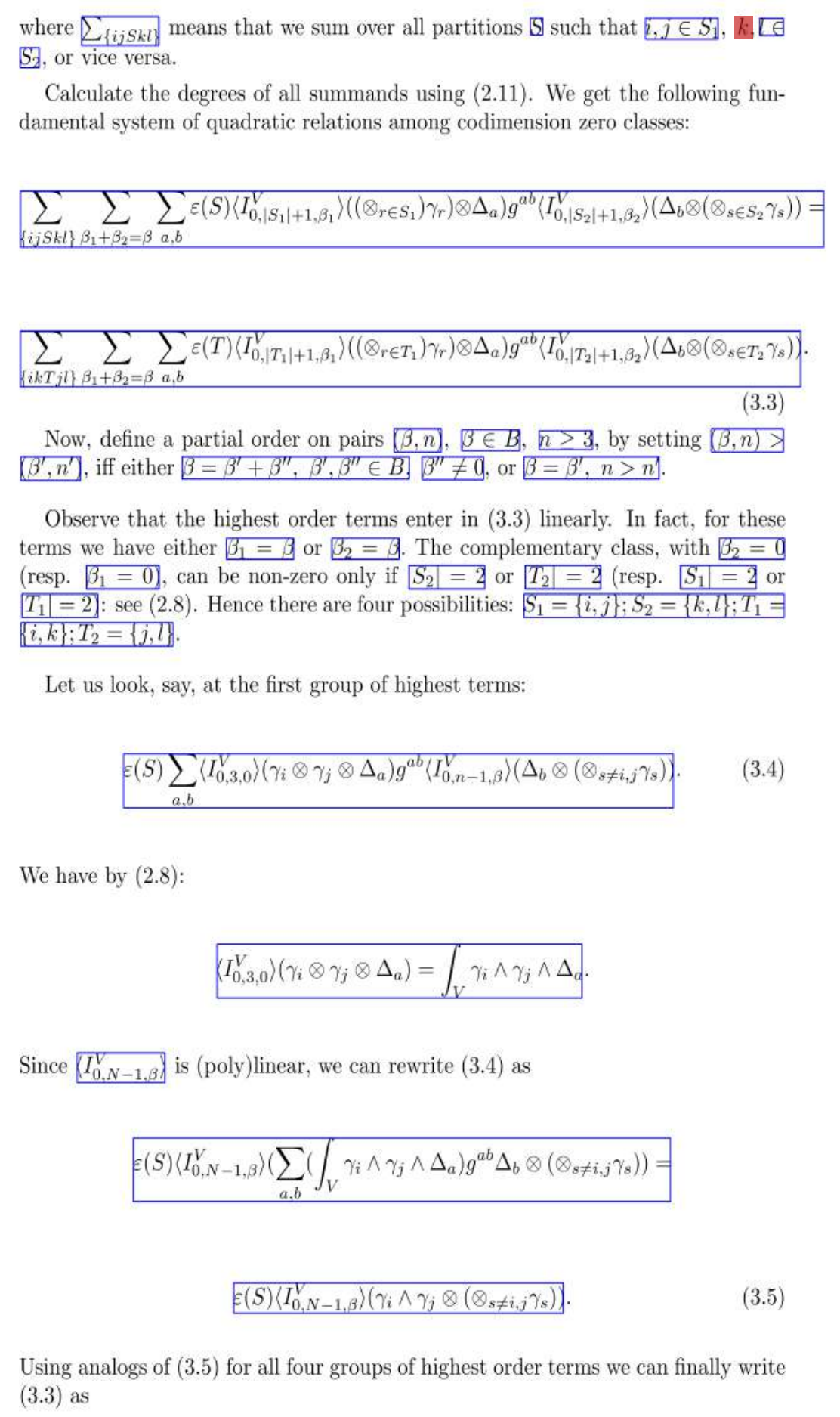}
    \hspace{-0.2in}
    &
		\includegraphics[width=0.23\textwidth,fbox]{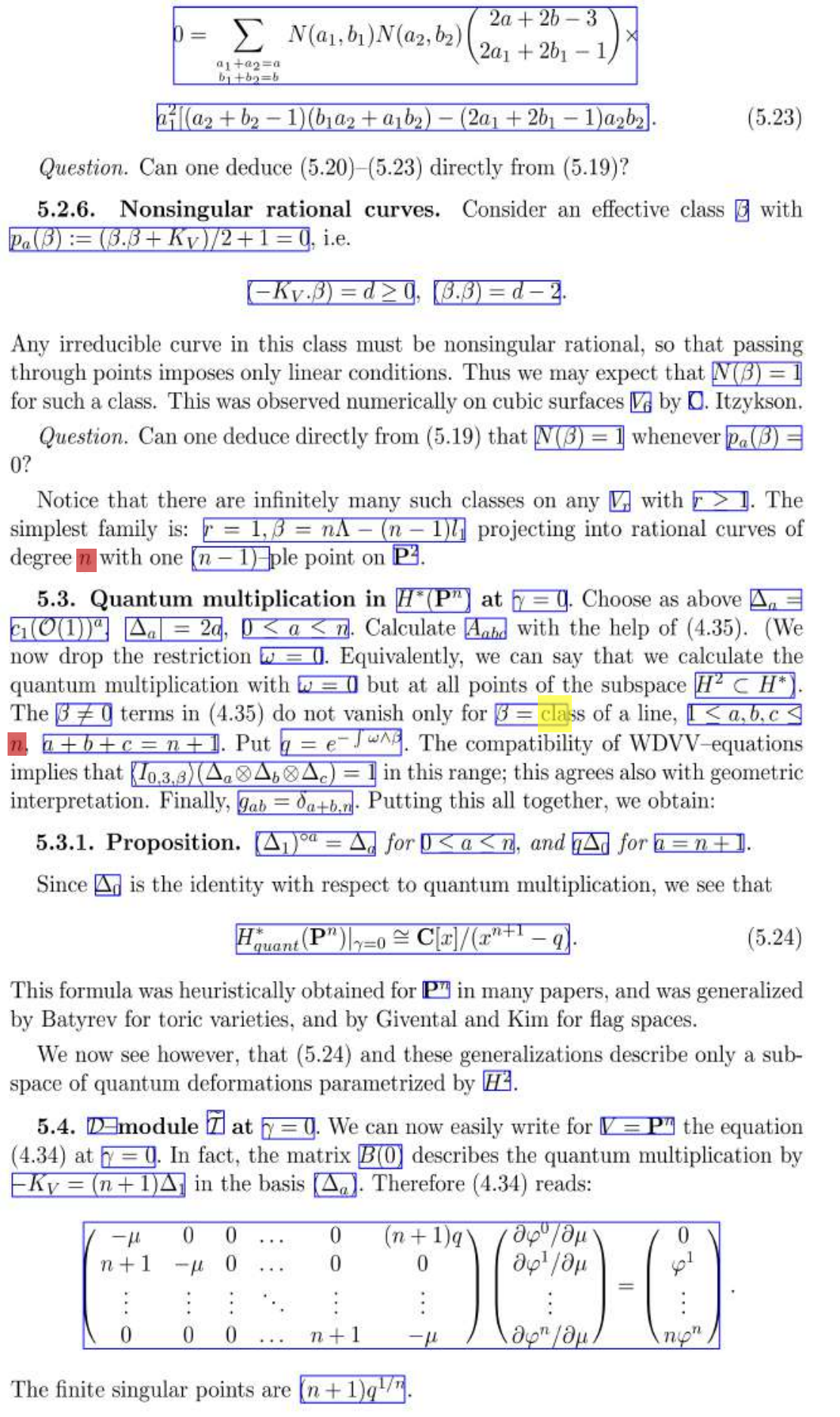}
     \hspace{-0.2in}
    &
	
		\includegraphics[width=0.23\textwidth, fbox]{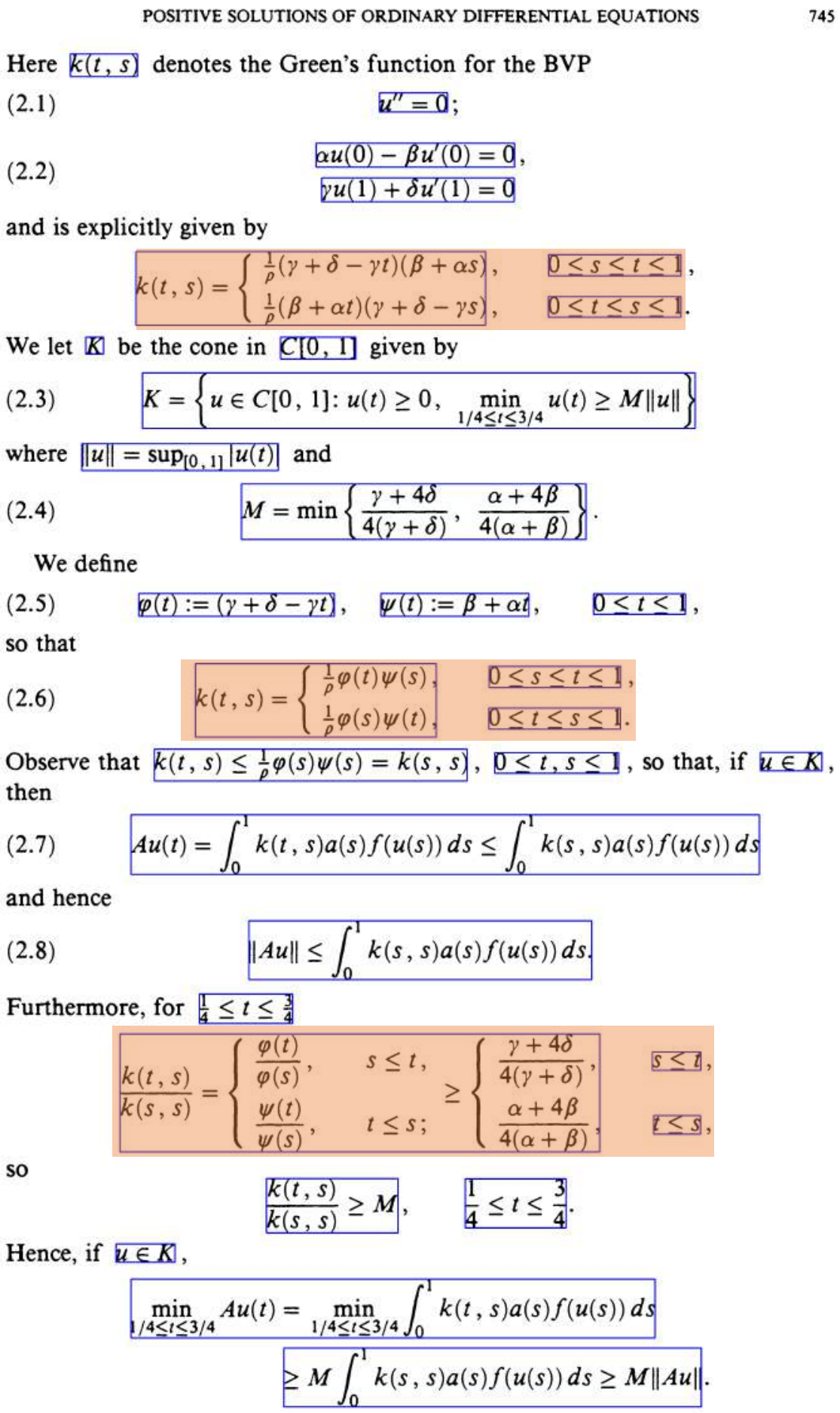}
	 \hspace{-0.2in}
	&
	    \includegraphics[width=0.23\textwidth, fbox]{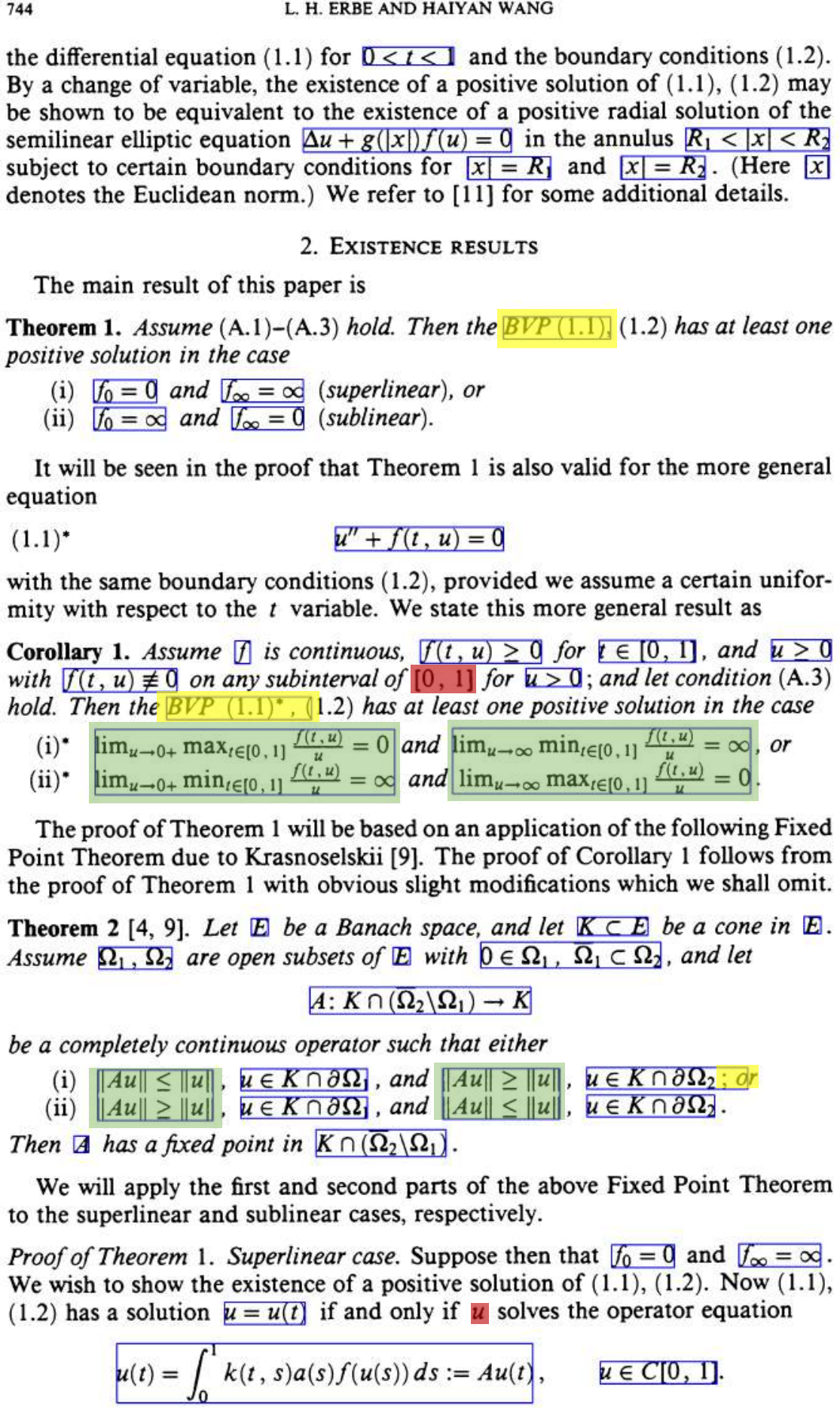}
	\end{tabular}
	
	\caption{Detection results. Detected formulas are shown as blue bounding boxes. Split formulas are highlighted in pink (3rd panel), and merged formulas are highlighted in green (4th panel). A small number of false negatives (red) and false positives (yellow) are produced.}
	\label{fig:det}
\end{figure*}

At inference time we send overlapping page windows to our modified SSD detector, and obtain formula bounding boxes with associated confidences in each window. As the SSD network sees the same page region multiple times, multiple bounding boxes are often predicted for a single formula (see Figure \ref{fig:detection_flow}).  Detections within windows are  stitched together on the page, and then each detection region votes at the pixel level.  Pixel-level votes are thresholded, and the bounding boxes of connected components in the resulting binary image are returned in the output as formula detections.  Example formula detection results are provided in Figure \ref{fig:det}.

{\bf Voting.} Let $B$ be the set of page-level bounding boxes for detected formulas, and $C$ be set of confidences obtained for each. Let $B_i\in B$ be the $i^{th}$ bounding box with confidence $C_i \in C$. Let each pixel in image $I$ be represented by pixel $P_{ab}$. We say that a pixel $P_{ab} \in B_i$ if it is inside the bounding box $B_i$. Let us define 
\begin{equation*}
L^i_{ab} = \begin{cases} 1 & \text{if}~P_{ab} \in B_i\\ 0 & \text{if}~P_{ab} \notin B_i \end{cases}
\end{equation*}

It is possible that $\sum_i L^i_{ab} \ge 1$, meaning that $P_{ab}$ belongs to more than one bounding box.  We considered different vote scoring functions $S_{ab}$ for each pixel $P_{ab}$:

\begin{quote}
\begin{tabular}{r l}
	 {\it uniform (count)} 
	& $\sum_{i=0}^{\mid B \mid} L^i_{ab}$\\~\\
	
	{\it max} 
	& $\argmax{i \in \{0, \ldots, |B|\}}{L^i_{ab} C_i}$\\~\\

	{\it sum} 
	&
	$\sum_{i=0}^{\mid B \mid} L^i_{ab} C_i$ \\~\\
	
	{\it average} 
	& $\sum_{i=0}^{\mid B \mid} L^i_{ab} C_i  ~/~  \sum_{i=0}^{\mid B \mid} L^i_{ab} $
	\\~\\
\end{tabular}
\end{quote}


{\bf Thresholding.}  We compare voting methods and tune their associated thresholds using the training data. A grid search was performed to maximize detection results for each voting method (f-score for IOU $\geq 0.75$). Average scoring does not perform as well as the other methods. For uniform weighting and sum scores, we tried thresholds in $\{ 0, 1, \ldots, 55 \}$, and for max scoring we tried thresholds in $\{ 0, 1, \ldots,100 \}$.  The simplest method where each pixel for the number of detections it belongs to (uniform weighting) obtained the best detection results using a threshold value of 30, and so we use this in our experiments. 


\section{Results and Discussion}
\label{sec:results}

\subsection{Training} 

We used a validation set to tune hyper-parameters for the ScanSSD detector. The TFD-ICDAR2019v2 training dataset was further divided into training (453 pages) and validation sets (116 pages). This produces 524,718 training and 131,999 testing sub-images, respectively. In our preliminary experiments, we observed that using a larger window size with SSD512 performs far better (+5\% f-score) than SSD300,\cite{paragthesis}  and cross-entropy loss with hard-negative mining performs better than focal-loss \cite{lin2017focal} (with or without hard-negative mining). Focal-loss reshapes the standard cross entropy loss such that it down-weights the loss for well-classified examples. 


We evaluated SSD models with different parameters\footnote{Details are available in \cite{paragthesis}}  and found that our HBOXES512 model, which introduces additional default box aspect ratios (see Section \ref{sec:matching_strategy}) performs better than SSD512, and MATH512 performs better than HBOXES512. For HBOXES512 we used  default boxes with aspect ratios $\{1,2,3,5,7,10\}$ instead of default boxes with aspect ratios $\{1, 2, 3, 1/2, 1/3\}$ for SSD512.   MATH512 uses default boxes with aspect ratios $\{1,2,3,5,7,10\}$ as well as rectangular kernels of size $1\times5$ rather than the square $3\times3$ kernel used in SSD512. From our experiments on the validation set, we observed that the MATH512 model consistently obtained the best detection results for the $512\times512$ inputs (by 0.5\% to 1.0\% f-score). So we use MATH512 for our evaluation. We then re-trained MATH512 using all TFD-ICDAR2019v2  training data. 

ScanSSD was built starting from an existing PyTorch SSD implementation.\footnote{\url{https://github.com/amdegroot/ssd.pytorch}} The VGG16 sub-network was pre-trained on ImageNet \cite{deng2009imagenet}. 

\subsection{Quantitative Results}

We used two evaluation methods, based on the ICDAR 2019 Typeset Formula Detection competition\cite{Mahdavi2019icdar2019} (Table \ref{table:results}), and the character-level detection metrics used by Ohyama et al.\cite{ohyama2019detecting} (Table \ref{table:benchmark}).

\begin{table}[!tb]
\caption{Results for TFD-ICDAR2019}
\scriptsize
\begin{tabular}{l|llllll}
\toprule
                         & \multicolumn{3}{c|}{IOU $\ge 0.75$}                & \multicolumn{3}{c}{IOU $\ge 0.5$} \\
                         & Precision & Recall & \multicolumn{1}{l|}{F-score} & Precision   & Recall  & F-score  \\ 
\midrule
ScanSSD\textsuperscript{*}\tnote{*}                  & \bf 0.781     & \bf 0.690  & \multicolumn{1}{l|}{\bf 0.733}   & \bf 0.848       & \bf 0.749   & \bf 0.796    \\

RIT 2\textsuperscript{$\dagger$}\tnote{***}                    & 0.753     & 0.625  & \multicolumn{1}{l|}{0.683}   & 0.831       & 0.670   & 0.754    \\
RIT 1                    & 0.632     & 0.582  & \multicolumn{1}{l|}{0.606}   & 0.744       & 0.685   & 0.713    \\
Mitchiking               & 0.191     & 0.139  & \multicolumn{1}{l|}{0.161}   & 0.369       & 0.270   & 0.312    \\ 
\hline
Samsung\textsuperscript{$\ddagger$}\tnote{**}             & 0.941     & 0.927  & \multicolumn{1}{l|}{0.934}   & 0.944       & 0.929   & 0.936    \\
\bottomrule
\end{tabular}

\begin{tablenotes}
     \item \textsuperscript{*} Used TFD-ICDAR2019v2 dataset
     \item \textsuperscript{$\dagger$} Earlier ScanSSD, placed $2^{nd}$ in TFD-ICDAR 2019 competition\cite{Mahdavi2019icdar2019}
     \item \textsuperscript{$\ddagger$} Used character information
\end{tablenotes}
\label{table:results}
\end{table}


{\bf Formula detection.}  An earlier version of ScanSSD placed second in the ICDAR 2019 competition on Typeset Formula Detection (TFD) \cite{Mahdavi2019icdar2019}.\footnote{The first place system used provided character information.}  The new ScanSSD system outperforms the other systems from the competition that did not use character locations and labels from ground truth. 

\begin{figure}
	\centering
	\includegraphics[width=0.5\textwidth]{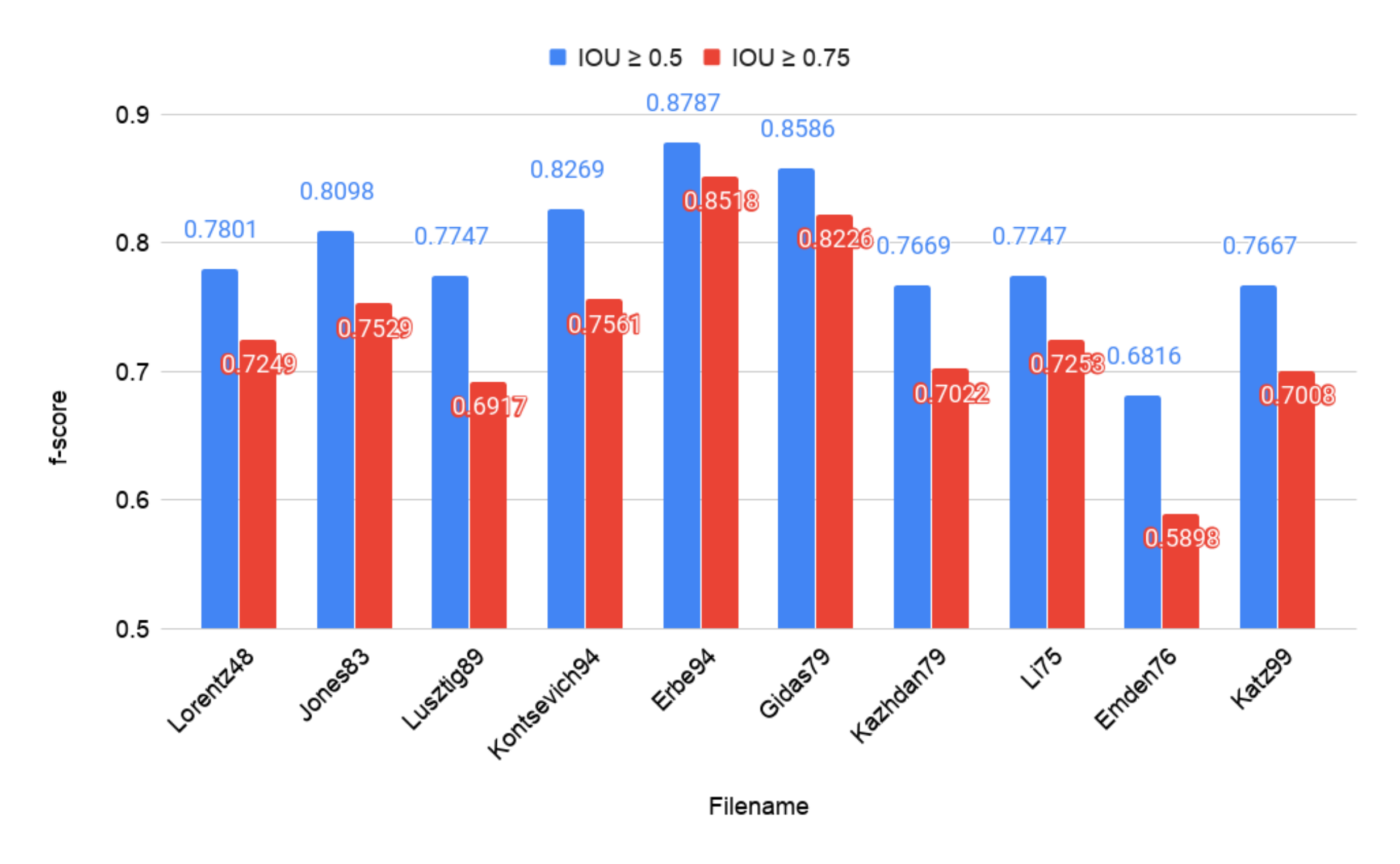}
	\caption{Document-level results, $IOU\geq0.5$ and $IOU\geq0.75$. }
	\label{fig:filewise}
\end{figure}

Figure \ref{fig:filewise} gives the document-level f-scores for each of the 10 testing documents, for matching constraints $IOU\geq0.5$ and $IOU\geq0.75$. The highest and lowest f-scores for $IOU\geq0.75$ are 0.8518 for Erbe94, and 0.5898 for Emden76. We think this variance is due to  document styles: we have more training documents with a style similar to Erbe94 than Emden76. With more diverse training data we expect better results. 

Examining the effect of the IOU matching threshold on results demonstrates that the detection regions found by ScanSSD are highly precise: 70.9\% of the ground-truth formulas are found at their exact location (i.e., IOU threshold of 1.0).  Requiring this exact matching of detected and ground truth formulas also yields a precision of 62.67\%, and an f-score of 66.5\%. To obtain a more complete picture, we next look at the detection of math symbols.

{\bf Math symbol detection.}
To measure math detection at the symbol (character) level, we consider all characters located within formula detections as `math' characters. Our method has 0.9652 recall and 0.889 precision at the character level, resulting in a 0.925 f-score.
This benchmarks well against recent results on the GTDB dataset (see Table \ref{table:benchmark}). Note that the detection targets (formulas for ScanSSD vs. characters), datasets, and evaluation protocols are different (1000 regions per test page are randomly sampled in Ohayama et al. \cite{ohyama2019detecting}), and so the measures are not directly comparable. The lower precision for character detection in ScanSSD may be an artifact of predicting formulas rather than individual characters.

The difference betweeen ScanSSD's math symbol detection f-score and formula detection f-score is primarily due to merging and splitting formula regions, which themselves are often valid subexpressions. Merging and splitting valid formula regions often produces regions too large or too small to satisfy the IOU matching criteria, leading to lower scores. Merging occurs in part because formula detections in neighboring text lines may overlap, and splitting may occur because large formulas have features similar to separate formulas within  windowed sub-images.


\begin{savenotes}
\begin{table}[!tb]
\centering
\caption{Benchmarking ScanSSD at the Character Level \cite{ohyama2019detecting}. Note differences in data sets and evaluation techniques (see main text).}
\resizebox{.4\textwidth}{!}{
\begin{tabular}{l|lll}
	\toprule
	\multicolumn{1}{c|}{\multirow{2}{*}{\textbf{System}}} & \multicolumn{3}{c}{\textbf{Math Symbol}} \\
	\multicolumn{1}{c|}{}                        & Precision     & Recall     & F-score     \\
	\midrule
	ScanSSD\textsuperscript{$\dagger$}\tnote{$\dagger$}                                      & 0.889         & 0.965      & 0.925       \\
	InftyReader\textsuperscript{*}\tnote{*}          & 0.971         & 0.946      & 0.958       \\
	ME U-Net\textsuperscript{*}                                     & 0.973         & 0.950      & 0.961\\
	\bottomrule      
\end{tabular}}

\begin{tablenotes}
     \item \textsuperscript{*} Used GTDB dataset
     \item \textsuperscript{$\dagger$} Used  TFD-ICDAR2019v2 dataset
\end{tablenotes}

\label{table:benchmark}
\end{table}
\end{savenotes}



\subsection{Qualitative results}

Figure \ref{fig:det} provides example ScanSSD detection results. ScanSSD can detect math regions of arbitrary size, from a single character to hundreds of characters. It also detects matrices and correctly rejects equation numbers, page numbers, and other numbers not belonging to formulas.  Figure \ref{fig:det} shows some example of detection errors. When there is a large space between characters within a formula (e.g., for variable constraints shown in the third panel of Figure \ref{fig:det}), ScanSSD may split the formula and generate multiple detections (shown with pink boxes). Second, when formulas are close to each other, our method may merge them (shown with  green boxes in Figure \ref{fig:det}). 
Another error not shown, was wide embedded graphs (visually similar to functions) being detected as math formulas.

On examination, it turns out that most detection `failures' are because of valid detections merged or split  in the manner described, and not spurious detections or false negatives. A small number of these are seen in Figure \ref{fig:det} using red and yellow boxes; note that all but one false negative are isolated symbols. 


\section{Conclusion}
\label{sec:conclusion}

In this paper we make two contributions:  \begin{enumerate*}
		\item modifying the GTDB datasets to compensate for differences in scale and translation found in the publicly available versions of PDFs in the collection, creating new bounding box annotations for math expressions, and
		
		\item the ScanSSD architecture for detecting math expressions in document images without using page layout, font, or character information. The method is simple but effective, applying a Single-Shot Detector (SSD) using a sliding window, followed by voting-based pooling across windows and scales. 
	\end{enumerate*}

Through our experiments, we observed that \begin{enumerate*}
		\item carefully selected default boxes improves formula detection,
		\item kernels of size $1 \times 5$ yield rectangular receptive fields that better-fit wide math expressions with larger aspect ratios, and avoid noise that square-shaped receptive fields introduce.	
	\end{enumerate*}


A key difference between formula detection in typeset documents and object detection in natural scenes is that typeset documents avoid occlusion of content by design. This constraint may help us design a better algorithm for non-maximal suppression, as the original non-maximal suppression algorithm is designed to handle overlapping objects.  Also, we would like to use a modified version of the pooling methods based on agglomerative clustering such as the fusion algorithm introduced by Yu et al.\cite{yu2018fusion}. We believe improved pooling will reduce the number of over-merged and split detections, improving both precision and recall. 

In our current architecture, we use a fixed pooling method; we plan to design an architecture where we can train the model end-to-end to learn pooling parameters directly from data. ScanSSD allows the use of multiple classes, and we would also like to explore detecting multiple page objects in a
 single framework. 
~\\

\noindent{\bf Acknowledgements.} This material is based upon work supported by the Alfred P. Sloan Foundation under Grant No. G-2017-9827 and the National Science Foundation (USA) under Grant No. IIS-1717997.

%
%



\bibliographystyle{IEEEtran}
\bibliography{IEEEabrv,main}

\end{document}